\documentclass[letterpaper, 10 pt, conference]{ieeeconf}  

\IEEEoverridecommandlockouts                              

\usepackage{booktabs}
\usepackage{multirow}
\usepackage{multicol}
\usepackage{graphicx}
\usepackage{hyperref}
\usepackage{xcolor}

\usepackage{caption}

\usepackage{background}
\newcommand{\mywatermark}{%
    \begin{minipage}{\textwidth}
        \centering
        \fontsize{10}{10}\selectfont 
        This paper has been accepted at the 28th IEEE International Conference on Intelligent Transportation Systems (ITSC 2025)
    \end{minipage}%
}

\backgroundsetup{
    scale=1,
    color=gray,
    angle=0,
    opacity=0.5,
    position=current page.north,
    vshift=-1cm, 
    contents={\mywatermark}
}

\title{\LARGE \bf
SEPose: A Synthetic Event-based Human Pose Estimation Dataset for Pedestrian Monitoring
}

\author{Kaustav Chanda, Aayush Atul Verma, Arpitsinh Vaghela, Yezhou Yang, and Bharatesh Chakravarthi%
\\
Arizona State University
}

\begin{document}

\twocolumn[{
\renewcommand\twocolumn[1][]{#1}%
\maketitle
\begin{center}    
    \centering
    \includegraphics[width=0.85\linewidth]{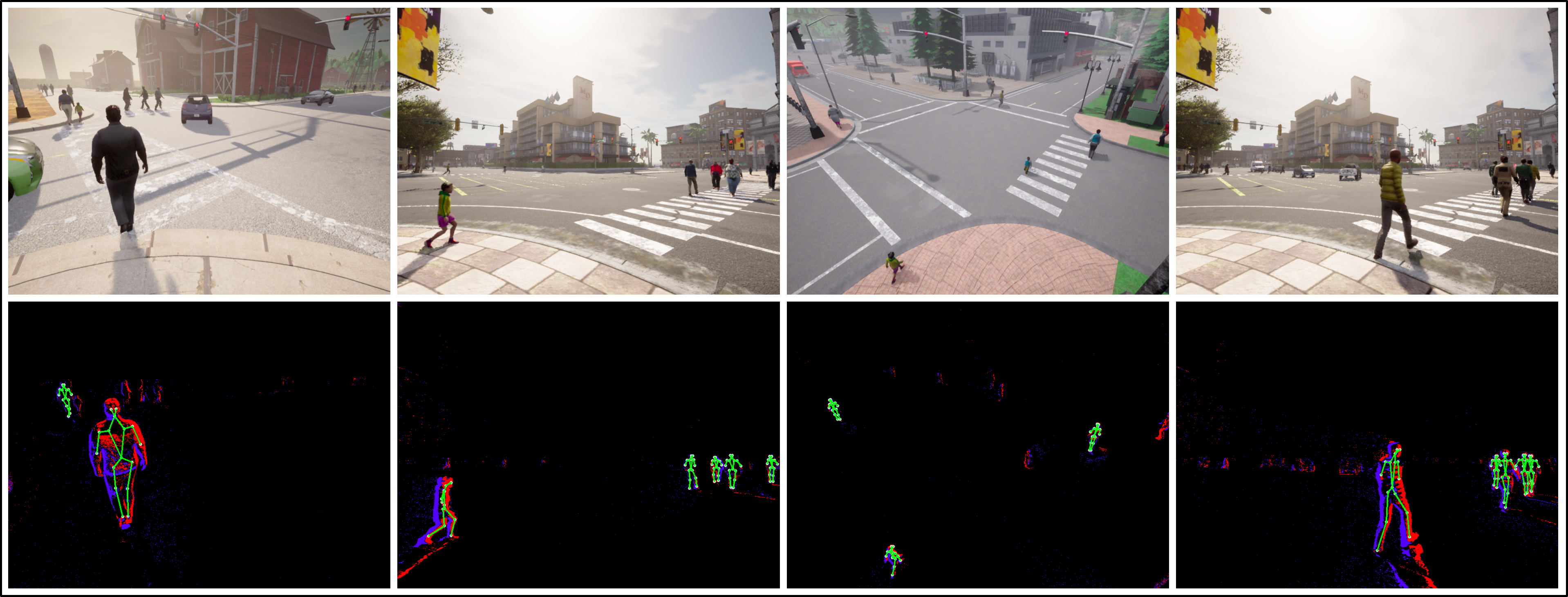}    
    \captionof{figure}{Visualization of SEPose data samples. (top row) RGB images captured by CARLA's virtual sensor and (bottom row) the corresponding event frames with overlaid ground-truth poses. }
    \label{fig_teaser}
\end{center}
}]

\makeatletter
\let\@fnsymbol\@arabic
\makeatother
\renewcommand{\thefootnote}{}
\footnotetext{This research is sponsored by NSF, the Partnerships for Innovation grant (\#2329780).}

\begin{abstract}
Event-based sensors have emerged as a promising solution for addressing challenging conditions in pedestrian and traffic monitoring systems. Their low-latency and high dynamic range allow for improved response time in safety-critical situations caused by distracted walking or other unusual movements. However, the availability of data covering such scenarios remains limited. To address this gap, we present SEPose -- a comprehensive synthetic event-based human pose estimation dataset for fixed pedestrian perception generated using dynamic vision sensors in the CARLA simulator. With nearly 350K annotated pedestrians with body pose keypoints from the perspective of fixed traffic cameras, SEPose is a comprehensive synthetic multi-person pose estimation dataset that spans busy and light crowds and traffic across diverse lighting and weather conditions in 4-way intersections in urban, suburban, and rural environments. We train existing state-of-the-art models such as RVT and YOLOv8 on our dataset and evaluate them on real event-based data to demonstrate the sim-to-real generalization capabilities of the proposed dataset.

\end{abstract}

\section{INTRODUCTION}

\begin{table*}[ht!]
\centering
\begin{tabular}{cc|c}
\toprule \midrule
\multicolumn{1}{c|}{\textbf{CARLA Map}} & \textbf{Description}                                                     & \textbf{No. of Samples} \\ \midrule \midrule
\multicolumn{1}{c|}{Town03}        & Large, urban map with a roundabout and large junctions                   & 13k                     \\ \hline
\multicolumn{1}{c|}{Town04}        & Small town embedded in the mountains                                     & 12k                     \\ \hline
\multicolumn{1}{c|}{Town05}        & Square-grid town with cross junctions and a bridge                      & 14k                     \\ \hline
\multicolumn{1}{c|}{Town07}        & Rural environment with narrow roads, corn and barns                      & 9k                      \\ \hline
\multicolumn{1}{c|}{Town10}        & Downtown urban environment with skyscrapers and residential buildings    & 13k                     \\ \hline
\multicolumn{1}{c|}{Town12}        & Large map with urban and rural regions. Data gathered from urban region. & 12k                     \\ \hline
\multicolumn{2}{c|}{\textbf{Total Number Of Frames}}                                                              & \textbf{73k}            \\ \midrule \bottomrule
\end{tabular}
\caption{Summary of the maps within the CARLA simulator used to generate SEPose data}
\label{map_desc}
\end{table*}

Despite technological advancements, modern traffic and pedestrian monitoring systems face critical challenges at the intersection of high vehicle speeds and the stringent requirement for near-instantaneous responses (low latency). High speeds necessitate rapid reaction times from intelligent traffic and pedestrian monitoring systems, which is often not possible with traditional methods. In this regard, event cameras \cite{delbruck_davis, chakravarthi2024recent} are a promising alternative. Inspired by the functioning of the human retina, these neuromorphic sensors capture asynchronous pixel-level changes in illumination with timestamps in the order of microseconds with an effective dynamic range of $120$$dB$. These characteristics carry the potential to be invaluable for addressing some of the aforementioned challenges.

In traffic monitoring applications, traditional frame-based cameras struggle on several fronts. One of them is speed, especially on highways \cite{chakravarthi2023event}. Considering the average speed of a vehicle to be around $70$ miles per hour, or about $31$ metres per second. With a frame rate of $30$ $fps$, the average highway vehicle would have moved about $1$ metre between two captured frames. This can potentially be significant in the context of road safety, which necessitates the use of sensing modalities that augment existing systems for faster response times. Event cameras carry the potential to revolutionize this field with their inherent properties of capturing spatially sparse but temporally dense data. An event stream is an array of $\left< x, y, p, t \right>$ tuples each indicating that the log intensity at the pixel location $\left< x, y, p,t \right>$ changed by a factor greater than a fixed threshold during the timestamp $t$. Thus, event cameras generate no data for static regions of the scene, leading to efficient data processing since dynamic objects can be localised directly. Although several event-based ego motion driving datasets like DDD17 \cite{binas2017ddd17endtoenddavisdriving}, DSEC \cite{gehrig2021dsecstereoeventcamera}, and MVSEC \cite{MVSEC} have been proposed, alongside event-based pose estimation datasets like PEDRo \cite{pedro} and DHP19 \cite{DHP19}, no large-scale event-based human pose estimation dataset focusing on pedestrian-traffic interactions has been proposed. This lack of data is the predominant reason for the limited adoption of event-based vision in the field of intelligent transportation systems (ITS). 

Additionally, a monitoring system utilizing event-based pedestrian pose data with the capacity to perform human pose estimation with low latency can serve as a platform for the development of systems that can raise timely alarms regarding pedestrian safety by predicting gait direction, detecting unusual movement patterns, and risk of collision with oncoming traffic. Such event-based systems have the potential to recognise unsafe situations and respond by taking immediate action, such as prolonging crossing time or alerting the concerned authorities. The availability of event-based pose estimation datasets is crucial for the development of predictive models for higher-level downstream tasks that require pose information, such as predicting future movement using the orientation of hip joints or the hand and torso positions for pedestrians in scooters or bicycles. Although frame-based datasets covering some of these tasks, such as the Waymo Open Dataset \cite{Ettinger_2021_ICCV} have been proposed, gathering and labelling a large-scale dataset that meets these requirements can be a prohibitively expensive undertaking, which is further complicated by the limited adoption of event cameras in traffic monitoring systems. In an effort to address these data requirements, we introduce \textit{SEPose} -- a Synthetic Event-based human Pose estimation dataset for pedestrian monitoring. The core contributions of this paper can be summarized as follows:
\begin{enumerate}
    \item A first-of-its-kind synthetic event-based human pose estimation dataset comprising about $73$K event frames with corresponding event streams and annotated human pose keypoints gathered using the CARLA \cite{dosovitskiy2017carlaopenurbandriving} simulator 
    \item The first synthetic pedestrian dataset simulating traffic camera data with a specific focus on pedestrian-pedestrian and pedestrian-vehicle interactions in a wide diversity of lighting and weather conditions across rural, urban, and suburban settings.
    \item A comprehensive sim-to-real study to evaluate the capability of the proposed dataset to generalize to real-world event data.
\end{enumerate}
The rest of the paper is organized as follows: \autoref{overview} provides a general overview of the motivations behind this work and a brief description of the software and hardware resources utilized. \autoref{gen_method} describes the data generation, processing, and cleaning process in detail, \autoref{pose_annotation_method} outlines how the dataset is structured and formatted, and \autoref{data_stats} provides some visualizations of the high-level characteristics of SEPose. \autoref{experiments} covers the simulation-to-real experiments and a discussion of the results from testing models trained with SEPose on real event camera data.

\section{Related Work}
Several prominent large-scale image-based datasets \cite{che2019d2citylargescaledashcamvideo, KITTI_robotics, Ettinger_2021_ICCV} have been proposed to accelerate research in the field of pedestrian perception. Early dataset work \cite{early_peds} in this field catered to classical methods such as Histograms of Oriented Gradients or AdaBoost cascade \cite{adaboost_cascade} for pedestrian and vehicle detection. With the deep learning revolution, successively larger datasets targeting specific challenging use cases, such as autonomous driving \cite{che2019d2citylargescaledashcamvideo, cordts2016cityscapesdatasetsemanticurban, apolloscape, geyer2020a2d2audiautonomousdriving}, were proposed. Another parallel line of work dealt with incorporating novel and diverse sensing techniques such as depth perception offered by LIDAR and IMU sensors \cite{apolloscape, chakravarthi2022real, ryu2022angular} and, more recently, DSEC \cite{gehrig2021dsecstereoeventcamera} and MVSEC \cite{MVSEC}, incorporating event cameras. Several multimodal datasets combining RGB data along with depth and 3D annotations for pedestrians, such as STCrowd \cite{cong2022stcrowdmultimodaldatasetpedestrian} and vehicles \cite{aliminati2024sevd, verma2024etram} have also been proposed. 

In contrast to vehicle detection, human pose estimation in traffic scenes presents its own challenges since pedestrians are relatively small for sensors to detect, can be present in crowds, and can have various poses, including running, micromobility, etc. Most early works focus on detection and semantic segmentation, such as \cite{lim2014crowdsaliencydetectionglobal}, Cityscapes \cite{cordts2016cityscapesdatasetsemanticurban}, and Citypersons \cite{citypersons}. Gradually, with the increased interest in pose estimation, more comprehensive datasets like COCOPose \cite{coco_pose} and Human3.6M \cite{human36} were proposed. Several works also utilize embedded sensors and devices for human pose estimation, such as smart watches \cite{abbasi2023watchpedpedestriancrossingintention}, or other peripherals \cite{rasouli2024divingdeeperpedestrianbehavior, desai2024cyclecrashdatasetbicyclecollision}. Although recently the problem of $3D$ human pose estimation has also been approached using diffusion models \cite{di2pose} and graph-based approaches \cite{jeong2024multiagentlongterm3dhuman}, human pose estimation research utilizing event-based sensors is still in its infancy, with DHP19 \cite{DHP19} and PEDRo \cite{pedro} being some of the prominent datasets.

\begin{figure*}[h]
  \centering
    \includegraphics[width=0.85\linewidth]{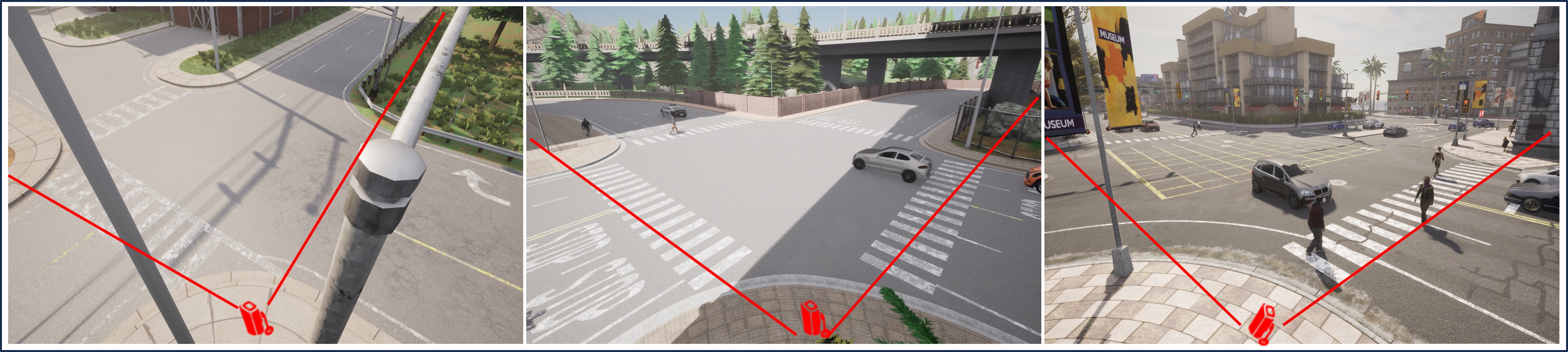}
    \caption{ Schematic representation of the camera positions in rural (left -- Town7), suburban (middle -- Town5), and urban (right -- Town10) settings. }
    \label{fig:cam_pos}  
\end{figure*}

\subsection{Event-Based Human Pose Estimation}
The DHP19 dataset addressed some of the challenges of high-speed human motion, such as motion blur that cripple traditional frame-based camera systems. Consisting of $18$ million frames with a resolution of $346 \times 260$ recorded using four synchronized DAVIS \cite{delbruck_davis} cameras in an indoor setup with subjects performing $33$ distinct movements such as hopping, punching, and kicking. The ground truth pose information is captured by a Vicon motion capture system, which tracks $13$ key joints in $3D$. Despite being an indoor-only dataset gathered in a controlled environment, the dynamic actions in which parts of the body are stationary present a challenging benchmark for event-based vision models, demanding robust temporal features. In contrast, PEDRo \cite{pedro} is an event-based dataset focused on multi-person detection in natural outdoor settings under varying weather conditions. It includes manually labelled bounding boxes recorded with a moving DAVIS camera. Although the outdoor setting and presence of multiple people in each frame make this dataset more useful for training pedestrian detection models, the absence of pose information and the eye-level point of view of the handheld camera restrict its utility. 

\section{SEPose overview}
\label{overview}

\begin{figure*}[ht!]
  \centering
    \includegraphics[width=0.85\linewidth]{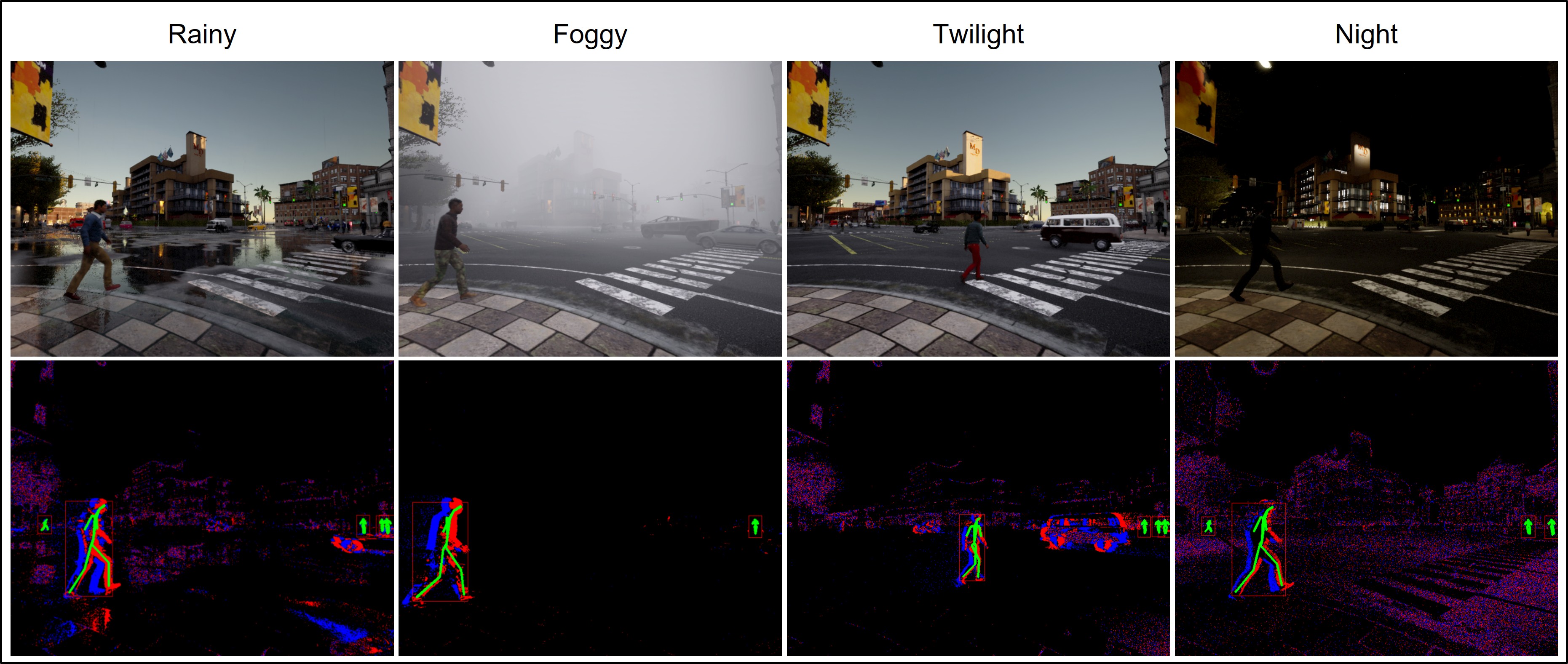}
    \caption{ Dataset samples of images from the RGB sensor (top) and DVS event camera (bottom) with overlaid human pose keypoints in rainy, foggy, twilight, and night time weather conditions. Blue pixels indicate positive event polarity, and red pixels indicate negative event polarity. }
    \label{fig:weathers}  
\end{figure*}

For the simulation environment, the CARLA \cite{dosovitskiy2017carlaopenurbandriving} simulator was chosen due to its realistic rendering engine built on top of Unreal Engine $4$ \cite{unrealengine}, and the flexibility offered by the API in generating custom scenarios. The simulator was run with the \textit{epic} rendering setting on a single NVIDIA TITAN X GPU. One of the key features of CARLA is the availability of $12$ highly detailed maps covering a wide range of commonly encountered driving scenarios, ranging from small rural towns with rivers and bridges, sub-urban settings with a mixture of residential and commercial buildings, to large downtown urban environments with skyscrapers and office buildings as shown in \autoref{fig_teaser}. The specific set of environments utilized for data generation is described in \autoref{map_desc}.

SEPose consists of RGB images, raw event stream in \textit{npz} format, event frames generated with an accumulation time of $33$$ms$, and corresponding annotations for the $2D$ human pose keypoints in the COCO \cite{coco_pose} format, along with other instance metadata such as time-of-day and weather.

\subsection{Sensor Configuration}

SEPose makes use of the RGB camera and the DVS event camera implemented in CARLA. Both the RGB as well as the DVS sensors have a field-of-view (FoV) of 90 degrees and resolution of 800 $\times$ 600 pixels. The DVS camera has a dynamic range of $140$ $dB$, and generates a continuous stream of events each of the form $\left< x, y, p, t \right>$ where an event with polarity $p \in \{-1,1\}$ is triggered at the pixel located at $(x,y)$ at timestamp $t$ when the log-intensity change $L$ exceeds a fixed threshold $p \times C$. In our simulation, $C$ is taken to be $0.3$ for both positive and negative events. The simulation is run using the synchronous mode to ensure that all of the sensor data is gathered at the same timestamp before the simulation is advanced. The simulation step time is kept at the default value of $33$$ms$.

\subsection{Data Generation Methodology}
\label{gen_method}

The data generation pipeline utilizes the pedestrian and vehicle behavior controllers implemented in CARLA. The main focus of the SEPose is to comprehensively simulate pedestrian behavior captured by standard traffic cameras in busy four-way intersections under realistic traffic conditions. For this purpose, the sensors are strategically placed in each map to either mimic standard traffic cameras placed at a height of about $10$$m$ or cameras placed adjacent to the driving lanes at approximately the height of a vehicle at about $2$$m$. The camera locations are approximately visualized in \autoref{fig:cam_pos}. The angle of the camera is chosen such that the crosswalks on either side are clearly visible in order to efficiently capture both pedestrian-pedestrian and pedestrian-vehicle interactions over a multitude of traffic signal cycles governed by a constant time traffic signal simulation implemented in CARLA. This is reflected in the heat map of the pedestrian bounding boxes visualized in \autoref{heatmap}. 

A key feature of SEPose is the diversity in both environments, as well as weather and lighting conditions. A wide range of weather conditions is covered by the proposed dataset- light to heavy rain, fog, and entire day-night time cycles. This is achieved by varying the weather parameters provided by the API, as shown in \autoref{weather_params}. Starting each from their default values, the weather parameters are sequentially incremented by 20\% every $100$ frames of the RGB camera. Data samples from a range of weather conditions are visualized in \autoref{fig:weathers}.

Due to the continuous nature of CARLA's traffic simulation, it is possible for the state of the world to reach undesirable states. For instance, since the traffic control algorithm does not take into consideration the size of other vehicles, they can sometimes collide, block other cars, or encroach upon sidewalks to avoid obstacles or stationary vehicles. This is particularly common if the initialization spawns an unusually high number of pedestrians or cars. During the data gathering phase, whenever a collision is detected, the last $100$ frames are discarded and the world is reloaded with the number of pedestrians and vehicles sampled from a Gaussian distribution as described in \autoref{pose_annotation_method}. This is done to ensure sufficient variety in both vehicle and pedestrian densities.

Two related problems critical to the validity of the simulated data are that of occlusions and the feasibility of pedestrian detection. Particularly, an effort is made to remove data samples where a significantly large proportion of the person is occluded by cars or other pedestrians. For this purpose, the ray-cast feature natively supported by UE4 is utilized. A ray is projected from the camera to each keypoint for every pedestrian in the scene. If greater than 50\% of the keypoints are occluded for a given pedestrian, then their corresponding entry is removed from the ground truth annotation file. Additionally, due to the nature of the scene being recorded, it is possible that some pedestrians may be far away from the camera to the extent that they're virtually indistinguishable. In such cases, although the ground truth keypoints are available, pedestrians whose bounding boxes are smaller than $300$$px$ in area are not recorded in the annotations.

\begin{table}[ht]
\centering
\begin{tabular}{c|c}
\toprule \midrule
\multicolumn{1}{c|}{\textbf{Parameter}} & \textbf{Range} \\ \midrule \midrule
\multicolumn{1}{c|}{cloudiness} &0 to 100 \\ \hline
\multicolumn{1}{c|}{precipitation} &0 to 100 \\ \hline
\multicolumn{1}{c|}{precipitation\_deposits} &0 to 100 \\ \hline
\multicolumn{1}{c|}{sun\_azimuth\_angle} &0 to 360\\ \hline
\multicolumn{1}{c|}{sun\_altitude\_angle} &-90 to 90\\ \hline
\multicolumn{1}{c|}{wetness} & $0$ - $100$ \\ \hline
\midrule \bottomrule
\end{tabular}
\caption{Summary of weather parameters covered by SEPose}
\label{weather_params}
\end{table}

\subsection{Pedestrian Pose Annotation}
\label{pose_annotation_method}

\begin{figure}[ht]
  \centering    
    \includegraphics[width=0.25\linewidth]{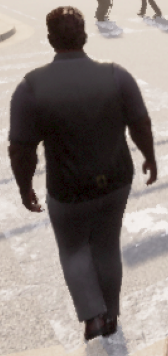}
    \includegraphics[width=0.25\linewidth]{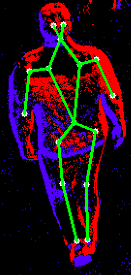}
    \caption{ (left) RGB image and (right) joint keypoints overlaid on the corresponding event frame.}
    \label{joints_vis}  
\end{figure}

The CARLA simulator implements a pedestrian controller that governs standard behavior, including walking along sidewalks, waiting for the signal before crossing, and avoiding oncoming vehicles as well as other pedestrians. The crowd density is constantly varied throughout our simulations, each time a new world is loaded, since the total number of pedestrians to spawn is sampled from a Gaussian distribution with a mean of $50$ and a standard deviation of $15$. This ensures sufficient variety in crowd density, ranging from light to heavily crowded scenes. For each pedestrian, we utilize the CARLA API's functionality to retrieve the ground truth skeleton in the origin frame. This is transformed into the frame of the RGB and DVS sensors, and the corresponding pixel coordinates are recorded in the COCO \cite{coco_pose} keypoint annotation format. The $16$ key points collected are visualized in \autoref{joints_vis}. For each RGB frame, the annotation file contains one line for each pedestrian with the ground truth joint keypoint locations. Each line is in the format $\left<C, x_1, y_1, x_2, y_2, ... x_{16}, y_{16} \right>$, where the joints are in the following order: 
\{ C\_Head, R\_eye, L\_eye, R\_shoulder, L\_shoulder, R\_arm, L\_arm, R\_foreArm, L\_foreArm,  C\_hips, R\_thigh, L\_thigh, R\_leg, L\_leg,  R\_foot, L\_foot \}, where C refers to the center, and R and L refer to the right and left counterparts of the corresponding joint.

\begin{table*}[ht]
    \centering    
    \begin{tabular}{lcc|cc|cc}
        \toprule \midrule
        \multirow{2}{*}{\textbf{Model}} & \multicolumn{2}{c|}{\textbf{SEPose}} & \multicolumn{2}{c|}{\textbf{PEDRo} \cite{pedro}} & \multicolumn{2}{c}{\textbf{DHP19} \cite{DHP19} } \\
        & \textbf{mAP} & \textbf{AP}$_{50}$ & \textbf{mAP} & \textbf{AP}$_{50}$ & \textbf{mAP} & \textbf{AP}$_{50}$ \\
        \midrule \midrule
        \textbf{RVT} & 69.9 & 98.1 & 44.4 (\textcolor{red}{36\%}) & 53.9 (\textcolor{red}{45\%}) & 51.1 (\textcolor{red}{27\%}) & 61.3 (\textcolor{red}{38\%}) \\
        \textbf{YOLO8X} & 63.2 & 94.7 & 39.4 (\textcolor{red}{37\%}) & 55.1 (\textcolor{red}{42\%}) & 53.7 (\textcolor{red}{15\%}) & 58.9 (\textcolor{red}{38\%}) \\
         \midrule \bottomrule
    \end{tabular}
    \caption{Summary of model performance when trained on SEPose and evaluated on PEDRo \cite{pedro} and DHP19 \cite{DHP19}. Percentage drops in the corresponding metrics are marked in red. }
    \label{table:sepose_eval}
\end{table*}

\subsection{Dataset Statistics}
\label{data_stats}

\begin{figure}[ht]
  \centering    
    \includegraphics[width=0.80\linewidth]{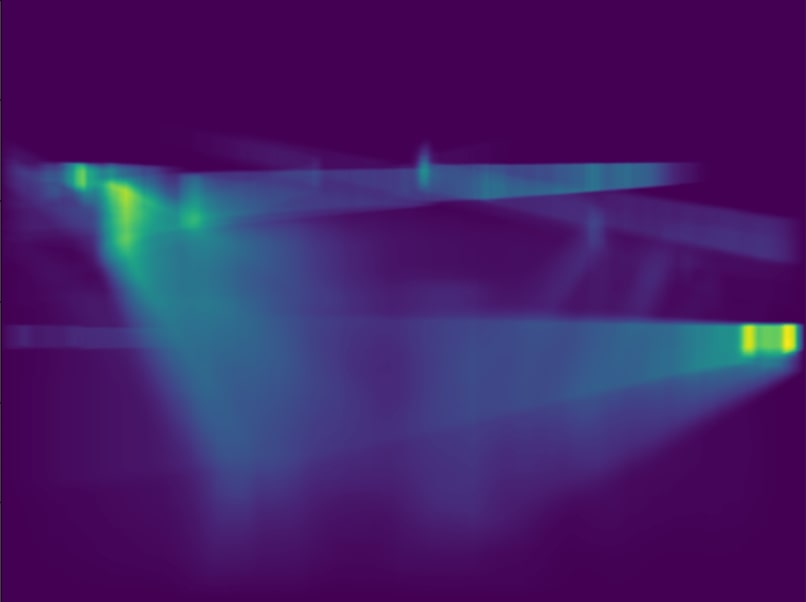}
    \caption{ Heatmap visualization of the regions of the frame covered by pedestrians across all maps in the dataset.}
    \label{heatmap}  
\end{figure}

\begin{figure}[ht]
  \centering    
    \includegraphics[width=0.96\linewidth]{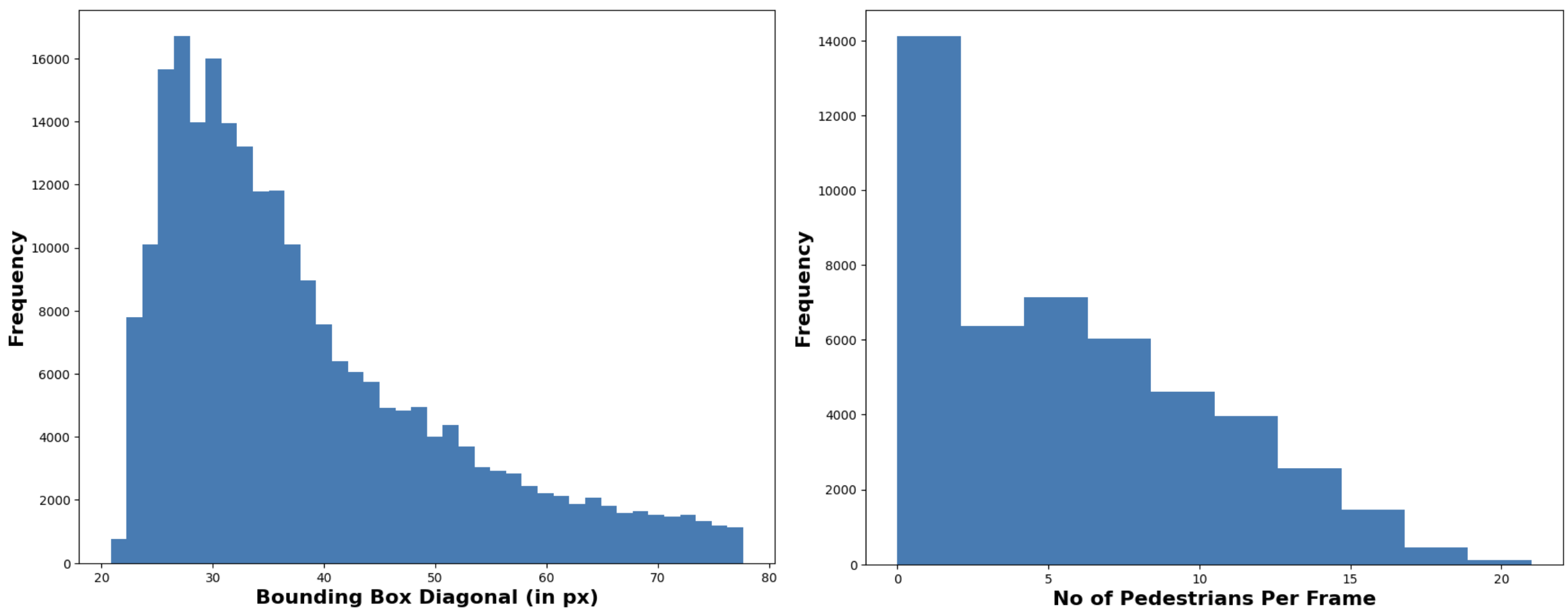}
    \caption{ (left) Histogram of bounding box diagonal size, (right) Histogram of the number of pedestrians in each frame after preprocessing with bin size of 2.}
    \label{bbox_stats}  
\end{figure}

In this section, we discuss the key characteristics of the proposed dataset. As discussed in \autoref{gen_method}, the influence of the chosen camera positions on the average location of pedestrians within each frame can be seen in \autoref{heatmap}. The four main lines correspond to the rough positions of the crosswalks across all of the maps in the dataset. It should be noted that the variation in the sizes of the bounding boxes can also be attributed to the variety in the pedestrian models, which may also include small children. This heatmap is in sharp contrast to existing indoor human pose estimation datasets, such as DHP19, in which the subjects are nearly always centered in the frame. Moreover, detecting pose keypoints where the subject is receding farther from the view of the camera is a challenging problem that is not encountered in indoor settings. The data cleaning methodology discussed in \autoref{gen_method} is reflected in both the histograms in \autoref{bbox_stats} for the bounding box diagonal size as well as the number of pedestrians per frame. 

With over $73$K frames and their corresponding event streams annotated with over $350$K human pose keypoints, SEPose is the largest synthetic pedestrian pose estimation dataset in the field of intelligent transportation systems. As a result of the consistent weather variations discussed in \autoref{gen_method}, the data samples are evenly distributed between all of the weather settings available in CARLA. This should also help models trained on SEPose to generalize to real-world event-based traffic data.

\section{Experiments}
\label{experiments}

A baseline is established for the performance gap between models trained on simulated and tested on real event camera data. We train current state-of-the-art model architectures such as the recurrent vision transformer (RVT) \cite{gehrig2023recurrentvisiontransformersobject} and Yolo8X \cite{ge2021yoloxexceedingyoloseries} on SEPose and test their performance on the real event camera datasets PEDRo \cite{pedro} and DHP19 \cite{DHP19}. Since SEPose includes both event streams as well as frames, both modalities are evaluated by training an RVT model with a Yolo detection head that takes event tensors as input, and a fully convolutional Yolo8X model is trained on event frames generated by the CARLA simulator with a time delta of 0.33 $ms$. The models are evaluated on DHP19 and PEDRo with their corresponding accuracies summarized in \autoref{table:sepose_eval}. DHP19 originally provides $3D$ keypoint annotations along with camera parameters to compute the $2D$ projections. This transformation is used to compute $2D$ keypoints for evaluation. As the annotated joints vary significantly across each of the three datasets compared, we generate bounding boxes from all pose annotations and compute the mean average precision and mAP50 scores for the corresponding intersection over union (IoU) scores. The percentage drop in accuracy for each corresponding metric shows the failure of the models to generalize to real-world data from the synthetic data generated by CARLA, although lower noise and the indoor setting may have contributed to a better performance for the DHP19 dataset compared to PEDRo, which includes more diverse indoor as well as outdoor scenarios. PEDRo is also a multi-person dataset in which the smaller bounding boxes for some subjects may have contributed to the drop in performance. Notably, in DHP19, all human subjects are centered in the image frame, while the SEPose dataset contains more side views and 3/4 view profiles. From \autoref{table:sepose_eval}, it is evident that although the performance drop is significant, models trained on synthetic data are able to learn meaningful features required for accurate predictions on real data. We hypothesize that simulating the noise model in real event cameras more accurately can help close this disparity in performance.

\section{Conclusion}
We introduce SEPose, a large, comprehensive event-based pedestrian pose estimation dataset specifically motivated towards advancing the state of the art in event-based research for intelligent transportation systems. At the intersection of human pose estimation and pedestrian monitoring, SEPose, being an event-based dataset, distinguishes itself from existing datasets that focus predominantly on detection or segmentation with traditional frame-based cameras. We hypothesize that widely available event-based human pose data can open up interesting avenues in intelligent pedestrian monitoring, enabling the potential for rapid response times and other applications, such as utilizing the hip angles to predict future direction of motion, which isn't generally possible with detection labels. To the best of our knowledge, a real-world counterpart of SEPose does not exist, which makes evaluation challenging since both DHP19 and PEDRo are not traffic-focused datasets. This can partly explain the significant drop in performance when models trained on SEPose are tested on them.


\bibliographystyle{ieeetr}
\bibliography{references}

\end{document}